\pdfoutput=1
\documentclass[11pt]{article}
\usepackage[]{acl}
\usepackage{times}
\usepackage{xurl}
\usepackage{hyperref}
\usepackage{amsmath}
\usepackage{latexsym}
\usepackage[T1]{fontenc}
\usepackage[utf8]{inputenc}
\usepackage{microtype}
\usepackage{booktabs}
\usepackage{multirow}
\usepackage{array}
\usepackage{graphicx}
\usepackage[export]{adjustbox}
\usepackage{tikz}
\usepackage{framed}
\usepackage{textcomp}

\usetikzlibrary{shapes.geometric, arrows, positioning, chains, shapes.callouts, backgrounds, shapes, arrows.meta, shadows}
\title{Empowering Air Travelers:\\A Chatbot for Canadian Air Passenger Rights}

\author{
Maksym Taranukhin$^1$~~~Sahithya Ravi$^{2,3}$~~~G\'abor Luk\'acs$^4$ \\
{\bf Evangelos Milios}$^1$~~~{\bf Vered Shwartz}$^{2,3}$ \\
$^1$ Faculty of Computer Science, Dalhousie University\\
$^2$ Department of Computer Science, University of British Columbia\\
$^3$ Vector Institute for AI\qquad
$^4$ Air Passenger Rights\\
\path{{m.t,eem}@cs.dal.ca, {sahiravi,vshwartz}@cs.ubc.ca,}\\
\path{lukacs@airpassengerrights.ca}
}

\begin{document}
\maketitle

%%%%%%%%%%%%%%%%%%%%%%%%%%%%%%%%%%%%%%%%%%%
\begin{abstract}
%%%%%%%%%%%%%%%%%%%%%%%%%%%%%%%%%%%%%%%%%%%
The Canadian air travel sector has seen a significant increase in flight delays, cancellations, and other issues concerning passenger rights. Recognizing this demand, we present a chatbot to assist passengers and educate them about their rights. Our system breaks a complex user input into simple queries which are used to retrieve information from a collection of documents detailing air travel regulations. The most relevant passages from these documents are presented along with links to the original documents and the generated queries, enabling users to dissect and leverage the information for their unique circumstances. The system successfully overcomes two predominant challenges: understanding complex user inputs, and delivering accurate answers, free of hallucinations, that passengers can rely on for making informed decisions. A user study comparing the chatbot to a Google search demonstrated the chatbot's usefulness and ease of use. Beyond the primary goal of providing accurate and timely information to air passengers regarding their rights, we hope that this system will also enable further research exploring the tradeoff between the user-friendly conversational interface of chatbots and the accuracy of retrieval systems.\footnote{The code is available at \url{https://github.com/maksym-taranukhin/apr_chatbot}}
\end{abstract}

%%%%%%%%%%%%%%%%%%%%%%%%%%%%%%%%%%%%%%%%%%%
\section{Introduction}
%%%%%%%%%%%%%%%%%%%%%%%%%%%%%%%%%%%%%%%%%%%
Air travel in Canada has seen many challenges when it comes to passenger rights. Canada's deficient regulations lag behind the standards adopted by other Western countries such as members of the European Union \cite{passengerProtectionReport}. Canada also lacks meaningful enforcement of passengers' existing rights by the federal regulator, whose cozy relationship with the airline industry and impartiality has been questioned by a Parliamentary committee \cite{covidIndustryImpactReport} and by the judiciary \cite{airpassengerVsCanada2021}. This situation has led to a high number of questions from passengers trying to understand their rights and find solutions. A group of dedicated volunteers\footnote{\url{https://airpassengerrights.ca}} is handling these questions, providing information on the rights and options available to affected passengers. However, the growing number of inquiries calls for a more efficient, automated solution to ensure quick and accurate responses.

\begin{figure}[t]
\centering
\includegraphics[width=.99\columnwidth]{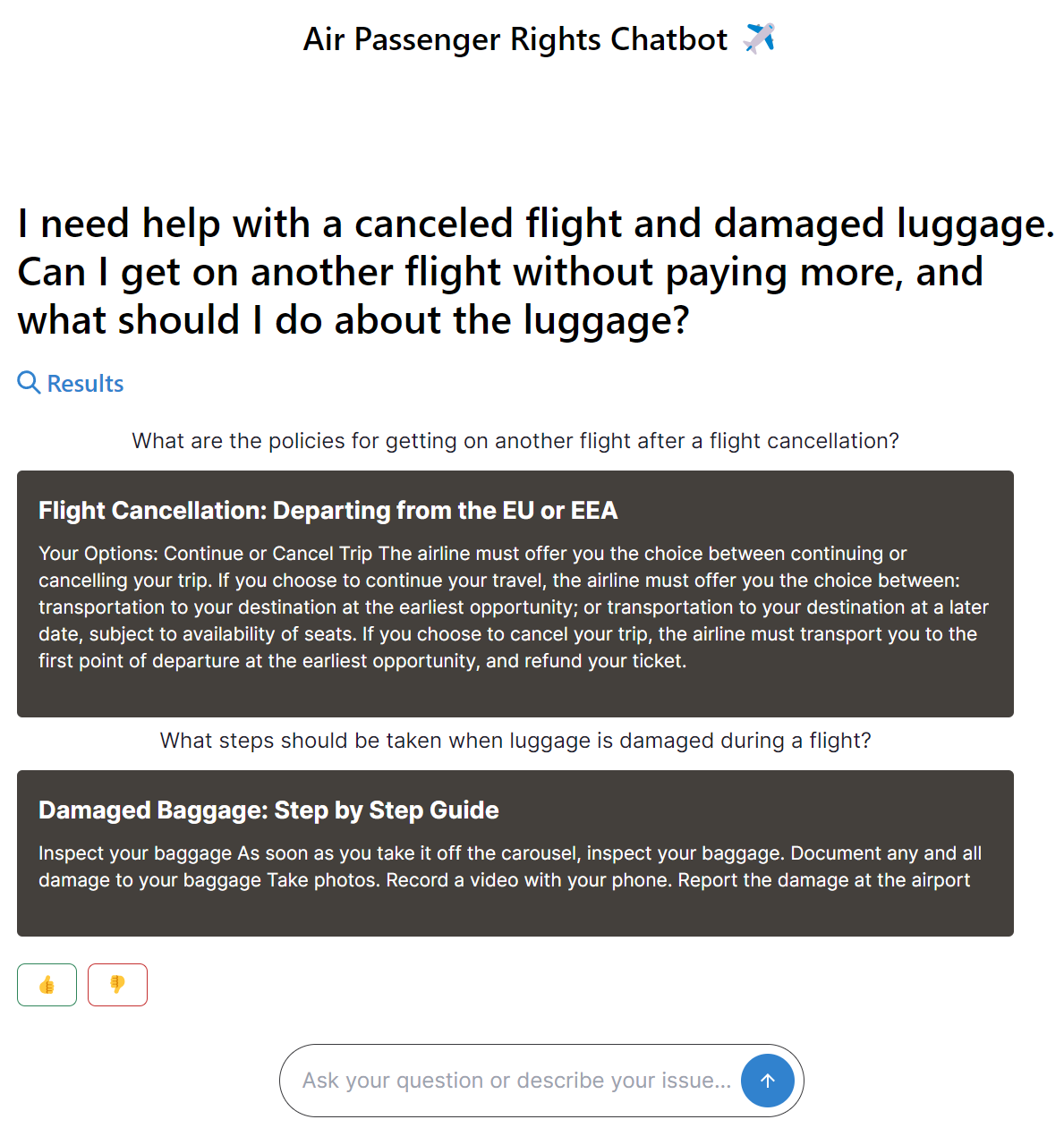}
\caption{User interface of the Air Passenger Rights chatbot.}
\label{fig:interface}
\end{figure}

To address this issue, we propose a chatbot (Figure~\ref{fig:interface}) that can adeptly understand narratives detailing air travel concerns and extract pertinent information from relevant sources. Our goal is to streamline the process of informing and educating passengers about their rights and options, ultimately empowering them to make informed decisions. This will reduce the workload of the human volunteers, allowing them to focus on the more complex questions from users.

Crucially, this application has a very low tolerance for errors and hallucinations, which may cost passengers time and money. Tellingly, in a recent incident, a chatbot developed by Air Canada provided incorrect information, leading to negative consequences for both the passenger and the airline.\footnote{\url{https://bc.ctvnews.ca/air-canada-s-chatbot-gave-a-b-c-man-the-wrong-information-now-the-airline-has-to-pay-for-the-mistake-1.6769454}}  
Our chatbot is designed to mitigate such risks via retrieval from a reliable collection of documents.  

Our approach simplifies complex questions, ensures systematic coverage of different aspects, and enhances search efficiency. 
Furthermore, to prevent 
hallucinations, we do not generate a response to the user based on the extracted information, as in the traditional RAG approach. Instead, we present the generated queries and the relevant passages from the source documents to the user. This method allows users to directly view the authoritative information that is relevant to their input, which they can then apply to their specific circumstances.

We conducted an extensive user study to evaluate the chatbot's performance across several dimensions: usefulness, user satisfaction, ease of use, and ease of learning. The results indicated that the chatbot was highly effective at providing pertinent information quickly and efficiently. The participants also reported that the chatbot's interface was more convenient than a manual Google search. Also, we compared our system to a standard RAG-based system and found that the latter had a hallucination rate of 27.5\%, which exceeds the acceptable threshold. In contrast, our chatbot produced zero hallucinations, highlighting its reliability in delivering accurate information.

In terms of the application itself, the proposed chatbot is first a prototype. Given the users' preference for the chatbot over a Google search, we are encouraged to develop future versions of the chatbot that are more conversational and that further contextualize the answers, while maintaining a strict zero-hallucination policy. The importance of this research goes beyond assisting air passengers; it introduces a way of using recent advances in NLP to provide legal information with greater accessibility and accuracy, especially in areas with complex regulations such as law and medicine.\footnote{Video demo is available \href{https://maksym-taranukhin.github.io/aprvideo.html}{here}.}

%%%%%%%%%%%%%%%%%%%%%%%%%%%%%%%%%%%%%%%%%%%
\section{Chatbot Architecture}
%%%%%%%%%%%%%%%%%%%%%%%%%%%%%%%%%%%%%%%%%%%
\begin{figure*}
    \centering
    \includegraphics[width=0.99\textwidth]{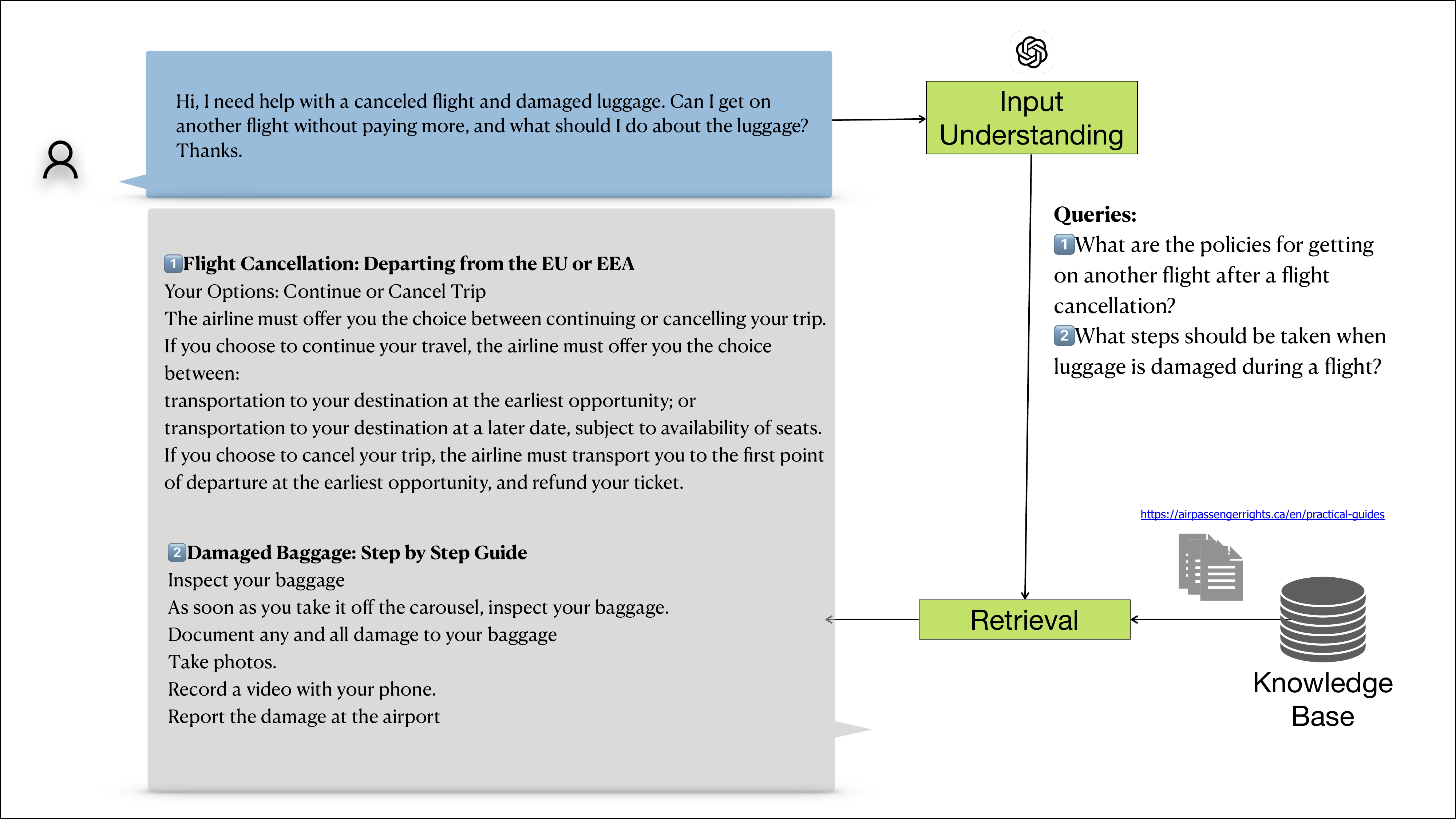}    
\caption{Overall Architecture of the Air Passenger Chatbot exemplified on an input query from the user: We use a LLM to decontextualize and decompose a given query, and provide a response by retrieving the relevant passages to answer the simplified queries.}
\label{fig:arch}
\end{figure*}

Our chatbot architecture, which is depicted in Figure~\ref{fig:arch}, is composed of 2 main components. The \emph{query understanding} component (Sec~\ref{sec:arch:query_understanding}) is responsible for interpreting the user input and generating a series of simpler queries. These queries go into the \emph{document retrieval} component (Sec~\ref{sec:arch:doc_retrieval}) which is tasked with extracting relevant information from the knowledge base. Finally, the extracted information is formatted and the answer is presented to the user (Sec~\ref{sec:arch:answer}).

\subsection{Input Understanding}
\label{sec:arch:query_understanding}

The query understanding component is specifically designed to handle complex and multi-part questions that require a nuanced understanding of context, intent, and the relationships between different pieces of information. In our chatbot, this component is built upon the GPT-4 model \cite{openai2023} and in-context learning to perform the following two key tasks. 

\paragraph{Decontextualization.}
Given a dialogue history and the current user input, the contextual query isolation component rephrases the current user input into a standalone text. For example, if the user input contains coreferences, such as referring to the previously mentioned airline using the pronoun ``they'', the contextual query isolation component will resolve these coreferences by replacing ``they'' with ``the airline''. This task can be considered a form of decontextualization in a dialogue context \cite{choi-etal-2021-decontextualization} and is crucial for ensuring that the user input can be understood and processed independently of its preceding conversation. See Appendix~\ref{sec:appx:prompts} for the prompt. 

\paragraph{Decompositional Query Generation.} Once isolated, the user input undergoes decompositional query generation where the goal is to dissect the standalone complex query into simpler, more manageable sub-questions. Consider the example in Figure~\ref{fig:arch}, where the user asks a question related to two distinct issues, namely, flight cancellation and damaged luggage. Decompositional query generation would parse this complex question into two simpler sub-questions focusing on each of the issues separately. Therefore, such division into discrete, more precise inquiries targets specific aspects of the original query, allowing for a more focused and efficient information retrieval process \cite{perez-etal-2020-unsupervised}. See Appendix~\ref{sec:appx:prompts} for the prompt. 

\subsection{Document Retrieval}
\label{sec:arch:doc_retrieval}

The document retrieval component is responsible for extracting relevant information from the knowledge base using the queries generated by the complex query understanding component. The document retrieval component employs a dense retrieval approach \cite{karpukhin-etal-2020-dense} to find the most relevant passages from the knowledge base for each generated query. Our dense retrieval approach uses OpenAI embeddings to encode both the queries and the documents into a high-dimensional space, and then uses cosine similarity to identify the top 5 relevant documents with scores greater than 0.7 for each query. We remove queries from the results if no relevant documents are found for them. 

\subsection{Answer Presentation}
\label{sec:arch:answer}

Once the relevant information is retrieved, the chatbot presents the information to the user in a structured manner as shown in Figure~\ref{fig:interface}. For each generated query, the chatbot provides the query and the corresponding passages from the source documents. This approach allows users to view the authoritative information that is relevant to their query, which they can then apply to their specific circumstances. By presenting the information in this way, the chatbot ensures that users receive accurate and reliable information, reducing the risk of model hallucinations that could occur if the system were to generate synthesized responses as in the traditional RAG architecture.

%%%%%%%%%%%%%%%%%%%%%%%%%%%%%%%%%%%%%%%%%%%
\section{Implementation Details}
%%%%%%%%%%%%%%%%%%%%%%%%%%%%%%%%%%%%%%%%%%%
\subsection{Data Collection}

Our chatbot utilizes a specialized knowledge base (KB), tailored specifically for addressing a variety of passenger issues in the Canadian air travel sector. This KB consists of domain-specific documents that extensively cover air travel regulations, with an emphasis on practical solutions for common problems such as flight delays, baggage mishandling, and boarding difficulties.

To construct this KB, we collected data from 88 web pages with regulatory details, step-by-step guidelines for resolving travel issues, a glossary of legal terminology, and other pertinent information. The documents were sourced primarily from two sections: the Practical Guides on the Air Passenger Rights website\footnote{\sloppy \url{https://airpassengerrights.ca/en/practical-guides}} and the Know Your Rights section from the Canadian Air Passenger Protection website.\footnote{\url{https://rppa-appr.ca/eng/know-your-rights}} We split all the documents (except step-by-step guides) by HTML headers to improve the precision of the retriever and to reduce information overload when the document is presented to the user.

\subsection{Web Application}

The chatbot is implemented as a web application with a backend built using Python and FastAPI, and a dynamic frontend created with Next.js, a popular React framework for developing web-based user interfaces. 

\paragraph{Backend.} The backend is responsible for the core functionalities of the chatbot, including processing user queries, extracting relevant information, and generating responses. The backend of our system leverages the GPT-4 model and OpenAI embeddings, accessed via the OpenAI API for the user input understanding and document retrieval components.\footnote{\url{https://platform.openai.com/docs/api-reference}} We set the generation temperature to 0 and the maximum sequence length to 300 tokens for GPT-4. The orchestration of all the backend components is managed using the LangChain library.\footnote{\url{https://python.langchain.com}}

\paragraph{Frontend.}
The frontend is a web-based interface that can be accessed from any device with an internet connection (Figure~\ref{fig:interface}). It is designed to be intuitive and user-friendly, allowing passengers to interact with the system easily. The interface includes a chat window where the user can type their queries and view the chatbot's responses. Each response from the chatbot contains the query text, relevant passages from the KB, and links to the source documents for users who want to explore the information in more detail.

%%%%%%%%%%%%%%%%%%%%%%%%%%%%%%%%%%%%%%%%%%%
\section{Evaluation}
%%%%%%%%%%%%%%%%%%%%%%%%%%%%%%%%%%%%%%%%%%%
\begin{table}[!htp]
\resizebox{\columnwidth}{!}{

\begin{tabular}{lcc}

\toprule
\textbf{Usability dimension} & \textbf{Chatbot} & \textbf{Google Search} \\
\midrule
Usefulness          & \textbf{1.75}     & 1.46 \\
Ease of use         & 1.06              & \textbf{1.23} \\
Ease of learning    & 2.18              & \textbf{2.27} \\
Satisfaction        & \textbf{2.46}     & 1.51 \\
\bottomrule

\end{tabular}
}

\caption{Usability test results. \textbf{Takeaways:} Users rated the chatbot as more useful and satisfying, while its ease of use and learnability were on par with Google Search.}
\label{tab:usability_results}

\end{table}
\begin{table*}[!htp]
\small
\centering
\begin{tabular}
{p{3.7cm}p{5.5cm}p{5.5cm}}
\toprule
\textbf{User Input} & \textbf{Chatbot} & \textbf{Google Search} \\
\midrule
\multirow{4}{3.7cm}{My \underline{flight was cancelled} and they \underline{lost my bag}. What are my compensation options?} &

\textbf{Query:} \textit{What are the compensation policies for flight cancellation?}
& 
\textbf{Query:} \textit{flight cancellations compensation} \\

& \textbf{Docs:}\newline
1. "Compensation for flight delays and cancellations"\newline
2. "Flight Cancellation General Principles"
& 
\textbf{Docs:}\newline
1. "Compensation for flight delays and cancellations" (5th place in search results)\\
\addlinespace

& \textbf{Query:} \textit{What are the compensation policies for lost luggage?}
& 
\textbf{Query:} \textit{lost luggage compensation }\\

& \textbf{Docs:}\newline
1. "Lost, damaged or delayed baggage"\newline
2. "Delayed Baggage: FAQ"
& 
\textbf{Docs:}\newline
1. "Delayed or Damaged Baggage (Air Canada)" (4th place in search results)
\\

\midrule
\multirow{2}{3.7cm}{Is there a \underline{time limit on filing} a luggage claim?} &
\textbf{Query:} \textit{Is there a deadline for filing a claim for lost luggage with an airline?} &
\textbf{Query:} \textit{lost luggage claim deadline} \\
&
\textbf{Docs:}\newline
1. "Filing a baggage claim with the airline"\newline
2. "Lost Baggage | General Principles" &
\textbf{Docs:}\newline
1. "Lost, damaged or delayed baggage" (2th place in search results)\\

\bottomrule
\end{tabular}
\caption{Comparative case study: Chatbot vs. Google Search for a compound travel issue. The retrieved documents are represented by their titles.}
\label{tab:case_study}
\end{table*}
\begin{table}[!htp]
\centering
\begin{tabular}{lcccc}
\toprule
& \textbf{P@5} & \textbf{R@5} & \textbf{F1@5} & \textbf{MAP@5} \\
\midrule
\textbf{Chatbot} & 0.78 & 0.83 & 0.8 & 0.88 \\
\bottomrule
\end{tabular}
\caption{Chatbot retrieval performance at top 5 documents.}
\label{tab:chatbot_retrieval_performance}
\end{table}

\subsection{User Study}

To evaluate the chatbot, we conducted a comparative usability study against a manual web search using Google. The test aimed to assess the chatbot's usefulness, user satisfaction as well as its ease of use and learning.

\paragraph{Methodology.} We recruited 15 participants who had no prior experience or familiarity with NLP technologies to ensure that the study outcomes were not influenced by the participants’ technical background. Each participant was asked to find information about 4 air travel-related issues randomly sampled from a pool of 40 issues covering a range of common passenger concerns, such as flight delays, cancellations, baggage issues and others. Each participant was asked to answer two questions using the chatbot and two others using Google search, with a random order of system used to control for any order effects.

Each session lasted approximately 30 minutes, during which participants interacted with each system to find answers for the assigned scenarios. To understand the user experience with each system, participants were asked to fill out a post-interaction survey immediately after using each system. The survey included both 7-point Likert scale questions (ranging from -3 for \textit{totally disagree} to +3 for \textit{totally agree}) measuring 4 usability dimensions based on the USE Questionnaire \cite{lund2001measuring}, as well as open-ended questions to collect free-form feedback such as opinions and suggestions. The questionnaire is available in Appendix~\ref{sec:appx:use_questionnaire}.

\paragraph{Quantitative Results.} Table~\ref{tab:usability_results} shows the average score for each usability dimension and each system. The chatbot scored notably high in terms of user satisfaction and usefulness. Out of a maximum of 3 points, the chatbot received an average score of 1.75 points for usefulness and 2.46 points for user satisfaction, with substantial gaps from the respective scores for Google search, especially for satisfaction. These findings suggest that the chatbot was more adept at providing targeted information quickly and effectively, leading to a more positive user experience. The chatbot scored close to Google search in terms of ease of use (-0.17 points difference) and ease of learning (-0.09 points difference). Given that participants are likely very well accustomed to searching Google, this suggests that the chatbot was intuitive to use.

\paragraph{Qualitative Results.} The participants' free-form feedback revealed that they appreciated the chatbot's conversational interface, which allowed for a more natural interaction. Some participants reported that they found the chatbot's direct answers to be more convenient than sifting through multiple search results on Google. On the downside, a few participants mentioned that the chatbot sometimes did not understand their queries or provided generic responses, which required rephrasing queries or formulating follow-up queries to get the desired information.

\subsection{Hallucination Analysis: Chatbot vs. RAG Approach}

Recent studies have shown that LLMs, such as GPT-4, are prone to generating responses that are inconsistent with legal facts in at least 58\% of cases for certain NLP tasks \cite{dahlLargeLegalFictions2024}. To assess how effectively the chatbot approach mitigates this issue, we conducted a quantitative comparison between the chatbot and the traditional RAG approach, focusing on answer hallucination. To this end, we manually evaluated the accuracy of the RAG system's response generation component using 40 examples from the user study, along with their corresponding ground truth documents, excluding the document retrieval component to avoid confounding factors. We measured the hallucination rate, defined as the percentage of responses containing information either not supported by the retrieved documents or factually incorrect.

The results showed that the RAG approach had a hallucination rate of 27.5\% (11 examples). Of these, 10\% (4 examples) were factually incorrect, while 22.5\% (9 examples) included information not present in the documents. These findings align with other studies in legal nlp, which reported hallucination rates between 17\% and 33\% for RAG-based systems \cite{hallucinationFree2024}. In contrast, the chatbot produces zero hallucinations, as it does not generate responses but instead presents the relevant passages from the source documents to the user.

Hallucinations can have serious consequences, particularly in high-stakes contexts where users rely on accurate information to make critical decisions. In the context of air travel regulations, even a minor hallucination could lead to a traveler misunderstanding the rules and facing delays or penalties. Hallucinations in the RAG approach can severely undermine user trust and lead to poor decision-making. Unlike the RAG approach, the chatbot's ability to completely avoid hallucinations makes it a more reliable tool for providing accurate information to users.

\subsection{Case Study}

We present a detailed case study to demonstrate the capability of the chatbot in handling complex air travel-related queries in comparison with manual Google searches, focusing on document relevance, interactivity, and efficiency. In Table~\ref{tab:case_study}, the dialogue showcases a scenario where a user seeks information on compensation for both a cancelled flight and lost luggage, followed by an inquiry about the time limit for filing a lost luggage claim.

\textbf{Document Retrieval.} The example highlights the chatbot's ability to directly retrieve top documents relevant to the user's queries. In contrast, the first relevant document for both flight cancellations and lost luggage appeared lower in Google search results (5th and 4th place, respectively). This highlights the chatbot's efficiency in swiftly providing relevant information to the user. Additionally, Table~\ref{tab:chatbot_retrieval_performance} provides quantitative results of the chatbot's performance in document retrieval at the top 5 documents as evaluated on 40 examples used in user-study, further confirming the chatbot's ability to prioritize the most relevant documents effectively.

\textbf{Interactivity.} The chatbot demonstrated superior interactivity by correctly interpreting a "luggage claim" as a claim for \textit{lost} luggage in the user's second turn. This ability to parse and respond to complex and context-dependent, multifaceted questions conversationally is a key advantage of dialogue systems over traditional search engines, which require users to input precise queries for each specific concern.

The effectiveness of the chatbot was evident in its ability to reduce the time and effort required from the user to obtain actionable information. Instead of navigating through multiple search results and possibly encountering irrelevant information (e.g., laws from other countries, news), the user received a tailored response that directly addressed their concerns.

%%%%%%%%%%%%%%%%%%%%%%%%%%%%%%%%%%%%%%%%%%%
\section{Related Work}
%%%%%%%%%%%%%%%%%%%%%%%%%%%%%%%%%%%%%%%%%%%
In recent years, research has focused on AI systems designed to aid individuals, especially those without legal expertise, in navigating complex legal procedures, bridging the gap between legal information and laypeople. The proposed chatbot operates within the domain of document-grounded dialogue systems (DGDS) that enable more trustworthy and informed user interactions. In this section, we overview the evolution of access to justice tools alongside the datasets and methods relevant to DGDS.

\subsection{Access to Justice Systems}
A variety of AI-driven systems have been developed to assist individuals without legal training in navigating legal processes, with a strong focus on addressing access to justice. Early systems used rule-based approaches to help litigants understand procedural requirements for specific legal domains, such as protection orders or housing issues \cite{10.1145/383535.383552, 10.1145/112646.112678}. Later systems expanded this by leveraging web-based, expert-guided platforms that further provided customized legal advice in areas like family law and consumer disputes \cite{thompson2015creating, bickel2015online}. More recent efforts have concentrated on hybrid systems that integrate rule-based reasoning with case-based analysis, enabling users to receive guidance based on both codified law and prior legal decisions \cite{10.1145/3594536.3595166, 10.1145/3322640.3326732}. Recent advancements in LLMs allow legal systems to scale across different domains without requiring extensive model training on vast amounts of data \cite{tan2023}. However, these models are prone to hallucination, generating plausible but factually incorrect legal advice, which could mislead users. To address this, we introduce a novel system that presents an answer to a user consisting of a set of extracted legal passages from a legal corpus, rather than generating a single response therefore improving the reliability of legal information and eliminating the risk of hallucination.

\subsection{Datasets for DGDS}
Incorporating documents into dialogue systems gained momentum with the rise of deep neural networks and large-scale datasets. One prominent dataset is the MultiWOZ \cite{budzianowski-etal-2018-multiwoz}, which comprises dialogues from a restaurant-search domain where the dialogue state is grounded in a set of documents containing information about hotels, restaurants, and other entities. Similarly,  \newcite{zhou-etal-2018-dataset} created a dataset with conversations based on Wikipedia articles about popular movies. In information-seeking DGDS, Doc2dial \cite{feng-etal-2020-doc2dial} and Multidoc2Dial \cite{feng-etal-2021-multidoc2dial} serve as realistic benchmarks to model goal-oriented information-seeking dialogues that are grounded on single or multiple documents.  An interesting data collection paradigm was investigated in QuAC \cite{choi-etal-2018-quac}, a Question Answering in Context dataset containing, 14000 information-seeking QA dialogues. The collection involved two crowd workers: one acting as to learn as much as possible about a hidden Wikipedia text, and one posing as a teacher who answers the questions by providing short excerpts from the text. 

\subsection{Approaches for DGDS}
In terms of approaches to DGDS, different methods for incorporation of external knowledge have been exhaustively explored to improve dialogue generation \cite{Lowe2015IncorporatingUT, liu-etal-2018-knowledge, Chen2019AWM, Sun2020HistoryAdaptionKI, Yu2020ASO}. A particular focus has been directed at knowledge selection, the process of choosing relevant contextual information \cite{Kim2020SequentialLK, Yang2022TAKETA, Sun2023GenerativeKS}. Some methods focus on the reasoning aspects of document-oriented dialogue, such as building an interpretable reasoning path to the evidence in the documents \cite{huang2018flowqa}, decomposing complex questions \cite{min-etal-2019-multi}, and performing multi-hop reasoning \cite{Tu2019SelectAA}. More recently, \cite{lai-etal-2023-external}, introduce a new
architecture for DGDS that includes a dense passage retriever, a re-ranker, and a response generation model. With the rise of LLMs as zero and few-shot learners, \newcite{Braunschweiler2023EvaluatingLL} perform a human evaluation as opposed to automatic evaluation of ChatGPT on document-grounded dialogue MultiDoc2Dial. In the context of the faithfulness of knowledge, \newcite{Razumovskaia2023textitDialBF} explore behavioral tuning to improve the faithfulness to the knowledge source in document-oriented dialogue.

%%%%%%%%%%%%%%%%%%%%%%%%%%%%%%%%%%%%%%%%%%%
\section{Conclusion}
%%%%%%%%%%%%%%%%%%%%%%%%%%%%%%%%%%%%%%%%%%%
We developed a chatbot that provides accurate and timely information about Canadian air travel regulations and passenger rights, supporting the manual process currently handled by a group of volunteers.

The chatbot utilizes retrieval augmented generation and in-context learning to interpret complex user inputs and extract relevant information from a comprehensive knowledge base. Instead of generating a synthesized response, it provides users with a direct presentation of the formulated queries and corresponding passages from the source documents, reducing the risk of hallucination. 

A user study comparing the chatbot to a Google search demonstrated its ability to accurately interpret and respond to user queries and successfully inform passengers of their rights. In future work, we plan to improve the chatbot's usefulness by contextualizing the answer for the user query, and reasoning over multiple extracted passages to synthesize an answer. We will explore how to achieve these properties without compromising the answers' accuracy.

%%%%%%%%%%%%%%%%%%%%%%%%%%%%%%%%%%%%%%%%%%
\section{Limitations}
%%%%%%%%%%%%%%%%%%%%%%%%%%%%%%%%%%%%%%%%%%
While our chatbot has shown promise in enhancing the accessibility of legal information regarding passenger rights, we recognize several limitations in the current iteration of the system that we plan to address in future work.

First, the chatbot's effectiveness is limited by its knowledge base's comprehensiveness. Missing information, like recent regulatory changes, can prevent it from providing complete answers. Therefore, it is crucial to continually expand and update the knowledge base to mitigate this limitation in a real system.

Secondly, the chatbot's current design does not facilitate an interactive dialogue which can be crucial for resolving uncertainties in user queries. For instance, if a user does not specify the origin and destination of their flight, the chatbot might not discern the applicable laws, as they can vary significantly from region to region—such as between Canada, Europe, and the United States. We plan to explore methods that allow the chatbot to ask follow-up questions to clarify such ambiguities.

Lastly, we've assumed users can understand and apply the legal information given, which might not hold true for everyone. Recognizing this, we intend to introduce simplified summaries and practical advice to enhance accessibility for users with varying levels of legal knowledge.

%%%%%%%%%%%%%%%%%%%%%%%%%%%%%%%%%%%%%%%%%%%
\section{Ethics Statement}
\paragraph{User Study.} Our user study scenarios are based on posts from the Air Passenger Rights (Canada) Facebook group.\footnote{\url{https://www.facebook.com/groups/441903102682254}} To protect user privacy, we anonymized the posts and used GPT-4 to generate variations covering a broad range of air travel issues, that we manually reviewed. We did not collect any personal information from the user study participants and we compensated participants CAD20 for a 30-minute session, which is well above the CAD16.75 hourly minimum wage in British Columbia, Canada.

\paragraph{User Privacy.} We used the paid API for GPT-4, which does not store user interactions, to respect user privacy and confidentiality. In future versions, we will consider switching to open-source locally-hosted LLMs instead. 

\paragraph{System Output.} Since our application has very little tolerance for providing users with the wrong information, we opted instead for a retrieval-based output. Thus, it is not subject to outputting offensive, dangerous, or factually incorrect text as do generative LLM-based models.
%%%%%%%%%%%%%%%%%%%%%%%%%%%%%%%%%%%%%%%%%%%

\section*{Acknowledgements}
We would like to express our sincere gratitude to the Digital Research Alliance of Canada for providing the computational resources that were instrumental in conducting the experiments and analysis presented in this paper.

Furthermore, we would like to acknowledge the invaluable contributions of Air Passenger Rights. The unique insights gained from this partnership greatly enriched the quality and impact of our research. Also, Vered's and Sahithya's research is supported by the Vector Institute for AI. Additionally Vered's research is supported by the CIFAR AI Chair program, and the Natural Sciences and Engineering Research Council of Canada. Evangelos' research is supported by the Natural Sciences and Engineering Research Council of Canada.

Finally, this research project has benefitted from the Microsoft Accelerate Foundation Models Research (AFMR) grant program.

\bibliography{custom, anthology}

\begin{thebibliography}{36}
\expandafter\ifx\csname natexlab\endcsname\relax\def\natexlab#1{#1}\fi

\bibitem[{{Air Passenger Rights, 2022}()}]{passengerProtectionReport}
{Air Passenger Rights, 2022}. 2022.
\newblock \href
  {https://www.ourcommons.ca/Content/Committee/441/TRAN/Brief/BR12177684/br-external/AirPassengerRights-e.pdf}
  {{From the Grounds Up: Revamp Canada's Air Passenger Protection Regime}}.
\newblock Accessed: October 24, 2023.

\bibitem[{Badawey(2021)}]{covidIndustryImpactReport}
Vance Badawey. 2021.
\newblock \href
  {https://www.ourcommons.ca/Content/Committee/432/TRAN/Reports/RP11431526/tranrp03/tranrp03-e.pdf}
  {{Emerging from the Crisis: A Study of the Impact of the COVID-19 Pandemic on
  the Air Transport Sector}}.
\newblock Report, Standing Committee on Transport, Infrastructure and
  Communities, House of Commons of Canada.
\newblock Accessed: 2023-11-19.

\bibitem[{Bickel et~al.(2015)Bickel, van Dijk, and Giebels}]{bickel2015online}
Esm{\'e}e~A Bickel, Maria Anna~Jozefa van Dijk, and Ellen Giebels. 2015.
\newblock \href
  {https://ris.utwente.nl/ws/portalfiles/portal/5136912/Online%20legal%20advice%20and%20conflict%20support_UTwente.pdf}
  {\emph{Online legal advice and conflict support: A Dutch experience}}.
\newblock University of Twente.

\bibitem[{Branting(2001)}]{10.1145/383535.383552}
L.~Karl Branting. 2001.
\newblock \href {https://doi.org/10.1145/383535.383552} {Advisory systems for
  pro se litigants}.
\newblock In \emph{Proceedings of the 8th International Conference on
  Artificial Intelligence and Law}, ICAIL '01, page 139–146, New York, NY,
  USA. Association for Computing Machinery.

\bibitem[{Braunschweiler et~al.(2023)Braunschweiler, Doddipatla, Keizer, and
  Stoyanchev}]{Braunschweiler2023EvaluatingLL}
Norbert Braunschweiler, Rama~Sanand Doddipatla, Simon Keizer, and Svetlana
  Stoyanchev. 2023.
\newblock \href {https://api.semanticscholar.org/CorpusID:262084173}
  {Evaluating large language models for document-grounded response generation
  in information-seeking dialogues}.
\newblock \emph{ArXiv}, abs/2309.11838.

\bibitem[{Budzianowski et~al.(2018)Budzianowski, Wen, Tseng, Casanueva, Ultes,
  Ramadan, and Ga{\v{s}}i{\'c}}]{budzianowski-etal-2018-multiwoz}
Pawe{\l} Budzianowski, Tsung-Hsien Wen, Bo-Hsiang Tseng, I{\~n}igo Casanueva,
  Stefan Ultes, Osman Ramadan, and Milica Ga{\v{s}}i{\'c}. 2018.
\newblock \href {https://doi.org/10.18653/v1/D18-1547} {{M}ulti{WOZ} - a
  large-scale multi-domain {W}izard-of-{O}z dataset for task-oriented dialogue
  modelling}.
\newblock In \emph{Proceedings of the 2018 Conference on Empirical Methods in
  Natural Language Processing}, pages 5016--5026, Brussels, Belgium.
  Association for Computational Linguistics.

\bibitem[{Chen et~al.(2019)Chen, Xu, and Xu}]{Chen2019AWM}
Xiuyi Chen, Jiaming Xu, and Bo~Xu. 2019.
\newblock \href {https://api.semanticscholar.org/CorpusID:196197298} {A working
  memory model for task-oriented dialog response generation}.
\newblock In \emph{Annual Meeting of the Association for Computational
  Linguistics}.

\bibitem[{Choi et~al.(2018)Choi, He, Iyyer, Yatskar, Yih, Choi, Liang, and
  Zettlemoyer}]{choi-etal-2018-quac}
Eunsol Choi, He~He, Mohit Iyyer, Mark Yatskar, Wen-tau Yih, Yejin Choi, Percy
  Liang, and Luke Zettlemoyer. 2018.
\newblock \href {https://doi.org/10.18653/v1/D18-1241} {{Q}u{AC}: Question
  answering in context}.
\newblock In \emph{Proceedings of the 2018 Conference on Empirical Methods in
  Natural Language Processing}, pages 2174--2184, Brussels, Belgium.
  Association for Computational Linguistics.

\bibitem[{Choi et~al.(2021)Choi, Palomaki, Lamm, Kwiatkowski, Das, and
  Collins}]{choi-etal-2021-decontextualization}
Eunsol Choi, Jennimaria Palomaki, Matthew Lamm, Tom Kwiatkowski, Dipanjan Das,
  and Michael Collins. 2021.
\newblock \href {https://doi.org/10.1162/tacl_a_00377} {Decontextualization:
  Making sentences stand-alone}.
\newblock \emph{Transactions of the Association for Computational Linguistics},
  9:447--461.

\bibitem[{Dahl et~al.(2024)Dahl, Magesh, Suzgun, and
  Ho}]{dahlLargeLegalFictions2024}
Matthew Dahl, Varun Magesh, Mirac Suzgun, and Daniel~E Ho. 2024.
\newblock \href {https://doi.org/10.1093/jla/laae003} {{Large Legal Fictions:
  Profiling Legal Hallucinations in Large Language Models}}.
\newblock \emph{Journal of Legal Analysis}, 16(1):64--93.

\bibitem[{{Federal Court of Appeal}, 2021()}]{airpassengerVsCanada2021}
{Federal Court of Appeal}, 2021. 2021.
\newblock \href
  {https://www.canlii.org/en/ca/fca/doc/2021/2021fca201/2021fca201.html\#par5}
  {{\itshape Air Passenger Rights v. Canada (Attorney General)}, {\rm 2021
  {fca} 201 at paras. 5-6}}.
\newblock Federal Court of Appeal.
\newblock 2021 FCA 201.

\bibitem[{Feng et~al.(2021)Feng, Patel, Wan, and
  Joshi}]{feng-etal-2021-multidoc2dial}
Song Feng, Siva~Sankalp Patel, Hui Wan, and Sachindra Joshi. 2021.
\newblock \href {https://doi.org/10.18653/v1/2021.emnlp-main.498}
  {{M}ulti{D}oc2{D}ial: Modeling dialogues grounded in multiple documents}.
\newblock In \emph{Proceedings of the 2021 Conference on Empirical Methods in
  Natural Language Processing}, pages 6162--6176, Online and Punta Cana,
  Dominican Republic. Association for Computational Linguistics.

\bibitem[{Feng et~al.(2020)Feng, Wan, Gunasekara, Patel, Joshi, and
  Lastras}]{feng-etal-2020-doc2dial}
Song Feng, Hui Wan, Chulaka Gunasekara, Siva Patel, Sachindra Joshi, and Luis
  Lastras. 2020.
\newblock \href {https://doi.org/10.18653/v1/2020.emnlp-main.652} {doc2dial: A
  goal-oriented document-grounded dialogue dataset}.
\newblock In \emph{Proceedings of the 2020 Conference on Empirical Methods in
  Natural Language Processing (EMNLP)}, pages 8118--8128, Online. Association
  for Computational Linguistics.

\bibitem[{Huang et~al.(2019)Huang, Choi, and tau Yih}]{huang2018flowqa}
Hsin-Yuan Huang, Eunsol Choi, and Wen tau Yih. 2019.
\newblock \href {https://openreview.net/forum?id=ByftGnR9KX} {Flow{QA}:
  Grasping flow in history for conversational machine comprehension}.
\newblock In \emph{International Conference on Learning Representations}.

\bibitem[{Karpukhin et~al.(2020)Karpukhin, Oguz, Min, Lewis, Wu, Edunov, Chen,
  and Yih}]{karpukhin-etal-2020-dense}
Vladimir Karpukhin, Barlas Oguz, Sewon Min, Patrick Lewis, Ledell Wu, Sergey
  Edunov, Danqi Chen, and Wen-tau Yih. 2020.
\newblock \href {https://doi.org/10.18653/v1/2020.emnlp-main.550} {Dense
  passage retrieval for open-domain question answering}.
\newblock In \emph{Proceedings of the 2020 Conference on Empirical Methods in
  Natural Language Processing (EMNLP)}, pages 6769--6781, Online. Association
  for Computational Linguistics.

\bibitem[{Kim et~al.(2020)Kim, Ahn, and Kim}]{Kim2020SequentialLK}
Byeongchang Kim, Jaewoo Ahn, and Gunhee Kim. 2020.
\newblock \href {https://api.semanticscholar.org/CorpusID:211146411}
  {Sequential latent knowledge selection for knowledge-grounded dialogue}.
\newblock \emph{ArXiv}, abs/2002.07510.

\bibitem[{Lai et~al.(2023)Lai, Castellucci, Kuzi, Ji, and
  Rokhlenko}]{lai-etal-2023-external}
Tuan~M. Lai, Giuseppe Castellucci, Saar Kuzi, Heng Ji, and Oleg Rokhlenko.
  2023.
\newblock \href {https://aclanthology.org/2023.eacl-main.264} {External
  knowledge acquisition for end-to-end document-oriented dialog systems}.
\newblock In \emph{Proceedings of the 17th Conference of the European Chapter
  of the Association for Computational Linguistics}, pages 3633--3647,
  Dubrovnik, Croatia. Association for Computational Linguistics.

\bibitem[{Liu et~al.(2018)Liu, Chen, Ren, Feng, Liu, and
  Yin}]{liu-etal-2018-knowledge}
Shuman Liu, Hongshen Chen, Zhaochun Ren, Yang Feng, Qun Liu, and Dawei Yin.
  2018.
\newblock \href {https://doi.org/10.18653/v1/P18-1138} {Knowledge diffusion for
  neural dialogue generation}.
\newblock In \emph{Proceedings of the 56th Annual Meeting of the Association
  for Computational Linguistics (Volume 1: Long Papers)}, pages 1489--1498,
  Melbourne, Australia. Association for Computational Linguistics.

\bibitem[{Lowe et~al.(2015)Lowe, Pow, Charlin, and
  Pineau}]{Lowe2015IncorporatingUT}
Ryan~Thomas Lowe, Nissan Pow, Laurent Charlin, and Joelle Pineau. 2015.
\newblock \href {https://api.semanticscholar.org/CorpusID:13812031}
  {Incorporating unstructured textual knowledge sources into neural dialogue
  systems}.
\newblock In \emph{Neural Information Processing Systems Workshop on Machine
  Learning for Spoken Language Understanding}.

\bibitem[{Lund(2001)}]{lund2001measuring}
Arnold~M Lund. 2001.
\newblock \href
  {https://www.researchgate.net/profile/Arnold-Lund/publication/230786746_Measuring_Usability_with_the_USE_Questionnaire/links/56e5a90e08ae98445c21561c/Measuring-Usability-with-the-USE-Questionnaire.pdf}
  {Measuring usability with the use questionnaire}.
\newblock \emph{Usability interface}, 8(2):3--6.

\bibitem[{Magesh et~al.(2024)Magesh, Surani, Dahl, Suzgun, Manning, and
  Ho}]{hallucinationFree2024}
Varun Magesh, Faiz Surani, Matthew Dahl, Mirac Suzgun, Christopher~D. Manning,
  and Daniel~E. Ho. 2024.
\newblock \href {http://arxiv.org/abs/2405.20362} {Hallucination-{{Free}}?
  {{Assessing}} the {{Reliability}} of {{Leading AI Legal Research Tools}}}.

\bibitem[{Min et~al.(2019)Min, Zhong, Zettlemoyer, and
  Hajishirzi}]{min-etal-2019-multi}
Sewon Min, Victor Zhong, Luke Zettlemoyer, and Hannaneh Hajishirzi. 2019.
\newblock \href {https://doi.org/10.18653/v1/P19-1613} {Multi-hop reading
  comprehension through question decomposition and rescoring}.
\newblock In \emph{Proceedings of the 57th Annual Meeting of the Association
  for Computational Linguistics}, pages 6097--6109, Florence, Italy.
  Association for Computational Linguistics.

\bibitem[{OpenAI(2023)}]{openai2023}
OpenAI. 2023.
\newblock \href {https://ar5iv.org/abs/2303.08774} {Gpt-4 technical report}.

\bibitem[{Paquin et~al.(1991)Paquin, Blanchard, and
  Thomasset}]{10.1145/112646.112678}
Louis-Claude Paquin, Fran\c{c}ois Blanchard, and Claude Thomasset. 1991.
\newblock \href {https://doi.org/10.1145/112646.112678} {Loge–expert: from a
  legal expert system to an information system for non-lawyers}.
\newblock In \emph{Proceedings of the 3rd International Conference on
  Artificial Intelligence and Law}, ICAIL '91, page 254–259, New York, NY,
  USA. Association for Computing Machinery.

\bibitem[{Perez et~al.(2020)Perez, Lewis, Yih, Cho, and
  Kiela}]{perez-etal-2020-unsupervised}
Ethan Perez, Patrick Lewis, Wen-tau Yih, Kyunghyun Cho, and Douwe Kiela. 2020.
\newblock \href {https://doi.org/10.18653/v1/2020.emnlp-main.713} {Unsupervised
  question decomposition for question answering}.
\newblock In \emph{Proceedings of the 2020 Conference on Empirical Methods in
  Natural Language Processing (EMNLP)}, pages 8864--8880, Online. Association
  for Computational Linguistics.

\bibitem[{Razumovskaia et~al.(2023)Razumovskaia, Vulic, Markovic, Cichy, Zheng,
  Wen, and Budzianowski}]{Razumovskaia2023textitDialBF}
Evgeniia Razumovskaia, Ivan Vulic, Pavle Markovic, Tomasz Cichy, Qian Zheng,
  Tsung-Hsien Wen, and Pawel Budzianowski. 2023.
\newblock \href {https://api.semanticscholar.org/CorpusID:265221328} {Dial
  beinfo for faithfulness: Improving factuality of information-seeking dialogue
  via behavioural fine-tuning}.
\newblock \emph{CoRR, abs/2311.09800}.

\bibitem[{Sun et~al.(2023)Sun, Ren, and Ren}]{Sun2023GenerativeKS}
Weiwei Sun, Pengjie Ren, and Zhaochun Ren. 2023.
\newblock \href {https://api.semanticscholar.org/CorpusID:258059716}
  {Generative knowledge selection for knowledge-grounded dialogues}.
\newblock In \emph{Findings}.

\bibitem[{Sun et~al.(2020)Sun, Hu, Xing, Yu, and
  Xie}]{Sun2020HistoryAdaptionKI}
Yajing Sun, Yue Hu, Luxi Xing, J.~Yu, and Yuqiang Xie. 2020.
\newblock \href {https://api.semanticscholar.org/CorpusID:213542942}
  {History-adaption knowledge incorporation mechanism for multi-turn dialogue
  system}.
\newblock In \emph{AAAI Conference on Artificial Intelligence}.

\bibitem[{Tan et~al.(2023)Tan, Westermann, and Benyekhlef}]{tan2023}
Jinzhe Tan, Hannes Westermann, and Karim Benyekhlef. 2023.
\newblock \href {https://ceur-ws.org/Vol-3435/short2.pdf} {Chatgpt as an
  artificial lawyer?}
\newblock In \emph{Artificial Intelligence for Access to Justice (AI4AJ 2023)}.

\bibitem[{Thompson(2015)}]{thompson2015creating}
Darin Thompson. 2015.
\newblock \href {https://ssrn.com/abstract=2696499} {Creating new pathways to
  justice using simple artificial intelligence and online dispute resolution}.
\newblock \emph{IJODR}, 2:4.

\bibitem[{Tu et~al.(2019)Tu, Huang, Wang, Huang, He, and Zhou}]{Tu2019SelectAA}
Ming Tu, Kevin Huang, Guangtao Wang, Jing Huang, Xiaodong He, and Bowen Zhou.
  2019.
\newblock \href {https://api.semanticscholar.org/CorpusID:207870753} {Select,
  answer and explain: Interpretable multi-hop reading comprehension over
  multiple documents}.
\newblock In \emph{AAAI Conference on Artificial Intelligence}.

\bibitem[{Westermann and Benyekhlef(2023)}]{10.1145/3594536.3595166}
Hannes Westermann and Karim Benyekhlef. 2023.
\newblock \href {https://doi.org/10.1145/3594536.3595166} {Justicebot: A
  methodology for building augmented intelligence tools for laypeople to
  increase access to justice}.
\newblock In \emph{Proceedings of the Nineteenth International Conference on
  Artificial Intelligence and Law}, ICAIL '23, page 351–360, New York, NY,
  USA. Association for Computing Machinery.

\bibitem[{Westermann et~al.(2019)Westermann, Walker, Ashley, and
  Benyekhlef}]{10.1145/3322640.3326732}
Hannes Westermann, Vern~R. Walker, Kevin~D. Ashley, and Karim Benyekhlef. 2019.
\newblock \href {https://doi.org/10.1145/3322640.3326732} {Using factors to
  predict and analyze landlord-tenant decisions to increase access to justice}.
\newblock In \emph{Proceedings of the Seventeenth International Conference on
  Artificial Intelligence and Law}, ICAIL '19, page 133–142, New York, NY,
  USA. Association for Computing Machinery.

\bibitem[{Yang et~al.(2022)Yang, Lin, Li, Meng, Wang, Wang, and
  Zhou}]{Yang2022TAKETA}
Chenxu Yang, Zheng Lin, JiangNan Li, Fandong Meng, Weiping Wang, Lan Wang, and
  Jie Zhou. 2022.
\newblock \href {https://api.semanticscholar.org/CorpusID:252819219} {Take:
  Topic-shift aware knowledge selection for dialogue generation}.
\newblock In \emph{International Conference on Computational Linguistics}.

\bibitem[{Yu et~al.(2020)Yu, Yu, Zhu, Li, Hu, Wang, Ji, and Jiang}]{Yu2020ASO}
W.~Yu, Wenhao Yu, Chenguang Zhu, Zaitang Li, Zhiting Hu, Qingyun Wang, Heng Ji,
  and Meng Jiang. 2020.
\newblock \href {https://api.semanticscholar.org/CorpusID:222272210} {A survey
  of knowledge-enhanced text generation}.
\newblock \emph{ACM Computing Surveys}, 54:1 -- 38.

\bibitem[{Zhou et~al.(2018)Zhou, Prabhumoye, and
  Black}]{zhou-etal-2018-dataset}
Kangyan Zhou, Shrimai Prabhumoye, and Alan~W Black. 2018.
\newblock \href {https://doi.org/10.18653/v1/D18-1076} {A dataset for document
  grounded conversations}.
\newblock In \emph{Proceedings of the 2018 Conference on Empirical Methods in
  Natural Language Processing}, pages 708--713, Brussels, Belgium. Association
  for Computational Linguistics.

\end{thebibliography}
\newpage

\appendix
\section{Prompts}
\label{sec:appx:prompts}

\subsection{Contextual Query Isolation Prompt}
\begin{framed}
\ttfamily \small \noindent
Given the following conversation and the follow-up input, rephrase the follow-up input into a standalone text that is not dependent on the conversation history. Make it as concise as possible, including only the necessary information.\\
\\
Chat History:\\
\{chat\_history\}\\\\
Follow Up Input:\\\{question\}\\\\
Text:
\end{framed}

\subsection{Decompositional Query Generation Prompt}
\begin{framed}
\ttfamily \small \noindent
Identify the information needed to respond to the following input. Provide your answer as a numbered list of questions, with each question focusing on a single, answerable aspect of the input. Limit the list to a maximum of 3 questions.\\
\\
Input: \{query\}\\\\
Questions:
\end{framed}

\section{USE Questionnaire}
\label{sec:appx:use_questionnaire}

The questionnaires were constructed as seven-point Likert rating scales, ranging from -3 (totally disagree) to +3 (totally agree)

\begin{enumerate}
    \item Usefulness
    \begin{enumerate}
        \item It helps me be more effective.
        \item It helps me be more productive.
        \item It is useful.
        \item It gives me more control over the activities in my life.
        \item It makes the things I want to accomplish easier to get done.
        \item It saves me time when I use it.
        \item It meets my needs.
        \item It does everything I would expect it to do.
    \end{enumerate}

    \item Ease of Use
    \begin{enumerate}
        \item It is easy to use.
        \item It is simple to use.
        \item It is user friendly.
        \item It requires the fewest steps possible to accomplish what I want to do with it.
        \item It is flexible.
        \item Using it is effortless.
        \item I can use it without written instructions.
        \item I don't notice any inconsistencies as I use it.
        \item Both occasional and regular users would like it.
        \item I can recover from mistakes quickly and easily.
        \item I can use it successfully every time.
    \end{enumerate}
    
    \item Ease of Learning
    \begin{enumerate}
        \item I learned to use it quickly.
        \item I easily remember how to use it.
        \item It is easy to learn to use it.
        \item I quickly became skillful with it.
    \end{enumerate}
    
    \item Satisfaction
    \begin{enumerate}
        \item I am satisfied with it.
        \item I would recommend it to a friend.
        \item It is fun to use.
        \item It works the way I want it to work.
        \item It is wonderful.
        \item I feel I need to have it.
        \item It is pleasant to use.
    \end{enumerate}
\end{enumerate}

\end{document}